\pdfoutput=1
\documentclass[letterpaper, 10 pt, conference]{ieeeconf}  

\IEEEoverridecommandlockouts                              

\usepackage{cite}
\usepackage{amsmath,amssymb,amsfonts}
\usepackage{algorithmic}
\usepackage{graphicx}
\usepackage{textcomp}
\usepackage{xcolor}
\usepackage{lipsum}
\usepackage{bm}
\usepackage{float}
\DeclareTextSymbolDefault{\DH}{T1}

\usepackage[utf8]{inputenc}
\usepackage{pgfplots}
\DeclareUnicodeCharacter{2212}{−}
\usepgfplotslibrary{groupplots,dateplot}
\usetikzlibrary{patterns,shapes.arrows}
\pgfplotsset{compat=newest}
\usepackage{subfigure}

\title{\LARGE \bf
Distributed Model Predictive Control for Cooperative Multirotor Landing on Uncrewed Surface Vessel in Waves 
}

\author{Jess Stephenson$^{*}$, Nathan T. Duncan$^{*}$ and Melissa Greeff 
\thanks{The authors are with Robora Lab (www.roboralab.com), Queen's University; and affiliated with Ingenuity Labs Research Institute. E-mails: jess.stephenson@queensu.ca, 20ntd1@queensu.ca, melissa.greeff@queensu.ca.}%
\thanks{*These authors contributed equally to this article.}
}

\begin{document}

\maketitle
\thispagestyle{empty}
\pagestyle{empty}

\begin{abstract}

Heterogeneous autonomous robot teams consisting of multirotor and uncrewed surface vessels (USVs) have the potential to enable various maritime applications, including advanced search-and-rescue operations. A critical requirement of these applications is the ability to land a multirotor on a USV for tasks such as recharging. This paper addresses the challenge of safely landing a multirotor on a cooperative USV in harsh open waters. To tackle this problem, we propose a novel sequential distributed model predictive control (MPC) scheme for cooperative multirotor-USV landing. Our approach combines standard tracking MPCs for the multirotor and USV with additional artificial intermediate goal locations. These artificial goals enable the robots to coordinate their cooperation without prior guidance. Each vehicle solves an individual optimization problem for both the artificial goal and an input that tracks it but only communicates the former to the other vehicle. The artificial goals are penalized by a suitable coupling cost. Furthermore, our proposed distributed MPC scheme utilizes a spatial-temporal wave model to coordinate in real-time a safer landing location and time the multirotor's landing to limit severe tilt of the USV.

\end{abstract}

\section{INTRODUCTION}

Leveraging a team of multirotor unmanned aerial vehicles (UAVs) and uncrewed surface vehicles (USVs) can significantly enhance maritime robotic applications, such as remote monitoring in the Arctic. UAVs provide high-speed aerial information, while USVs may be better suited for other tasks and offer extended range. In these applications, it is beneficial to land a multirotor on a USV for recharging.

Autonomous landing on a USV presents two main challenges. Firstly, it requires safe and reliable performance despite limited communication between the vehicles, and even in cases of temporary communication loss. Secondly, the USV may encounter rough water conditions, making the precise location and timing of the landing crucial to prevent damage due to the severe tilt of the USV during touchdown \cite{xia2022landing}.

\begin{figure}
\centering
\includegraphics[width=0.47\textwidth, trim={0cm 2cm 0 3cm},clip]{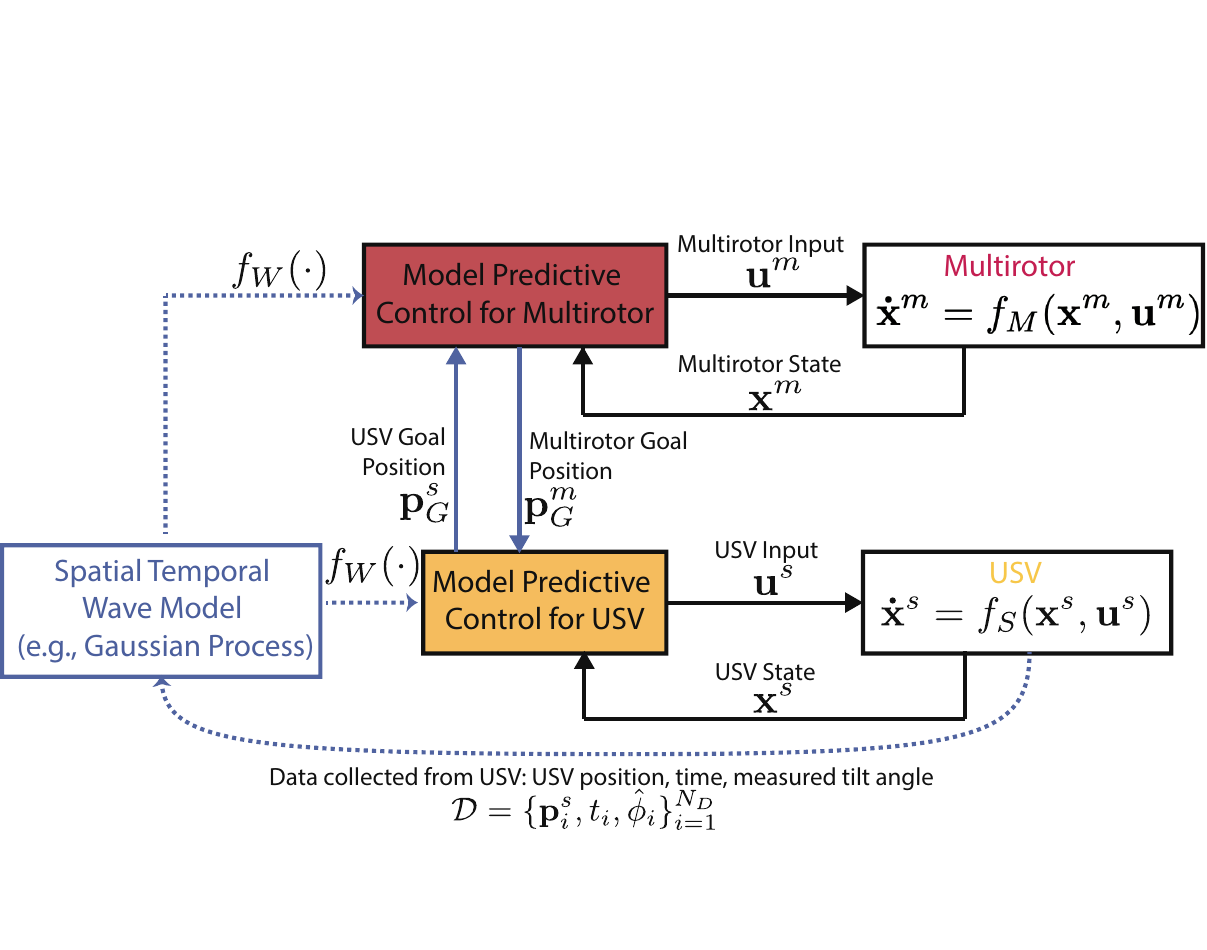}
\caption{Block diagram of our proposed distributed model predictive control (MPC): Our approach uses standard tracking MPCs for a multirotor and USV augmented with artificial goal locations. Each vehicle solves an individual optimization problem for both the artificial goal and an input that tracks it but only communicates the former. Our proposed distributed MPC simultaneously finds a consensus landing location between the UAV and USV (through  cooperation cost $J^{\text{co-op}}(\cdot)$), tracks it (through a tracking cost $J^{\text{track}}(\cdot)$), and leverages a spatial-temporal wave model $f_W(\cdot)$ to optimize a location and time that aids safe landing by minimizing large tilt angles of the USV (through a tilt cost $J^{\text{tilt}}(\cdot)$).}
\end{figure}

An optimization-based controller, such as Model Predictive Control (MPC), is one common strategy for multirotor landing tasks (e.g., \cite{gupta2022landing}, \cite{mohammadi2020vision}, \cite{pozzan2022non}) as it capitalizes on knowledge of the multirotor dynamical model while adhering to necessary safety constraints. 

One approach to address the landing task of heterogeneous UAV-USV agents is to use a centralized MPC (Model Predictive Control) \cite{persson2021model}. However, this approach has some limitations. It requires communication with a single centralized station, which makes it vulnerable to communication breaks and delays \cite{lapandic2021aperiodic}. Additionally, it involves solving a larger optimization problem that considers both vehicles' dynamics, which usually takes more time to solve. The alternative schemes are either decentralized or distributed control.

A decentralized MPC requires no communication between vehicles. A decentralized MPC has been demonstrated for landing a multirotor on a USV under the influence of waves \cite{gupta2022landing}. This is done by predicting the motion of the USV under waves using a camera onboard the multirotor. However, it assumes that the USV is stationary in space and is waiting for the multirotor to land while controlling its global positioning on the water. Instead, we adopt a distributed MPC architecture in this paper, where the USV can cooperate with the multirotor to select a suitable landing location. 

A distributed MPC architecture has been applied to landing a multirotor on a UGV \cite{garegnani2021autonomous} and USV \cite{bereza2020distributed}. The landing is treated as a rendezvous problem requiring both the multirotor and another vehicle to converge to a specific location in time. This rendezvous point is either known a priori or can be updated via an online heuristic. In this approach, the two vehicles share their planned or optimized trajectories.  

In our proposed approach, we treat the landing problem as a consensus problem where both vehicles have different goal locations and iteratively update their states and goals until they reach a consensus. The consensus location is not known a priori. The UAV and USV achieve self-organized consensus by leveraging a distributed MPC scheme similar to \cite{kohler2022distributed}. In \cite{kohler2022distributed}, a novel sequential distributed MPC scheme for nonlinear multi-agent systems is presented where each agent uses a local tracking MPC formulation with an artificial reference. No reference trajectory is supplied to the agents, and the final consensus goal is not given. Instead, the reference is implicitly provided by an additional cost that encodes the consensus goal. In this way, tracking and cooperative coordination are combined into one scheme. Non-cooperative fast multirotor landing on a moving UGV was achieved by leveraging vision-based localization of the platform and robust control techniques \cite{paris2020dynamic} \cite{baca2019autonomous}. This work considers a cooperative USV that aids in the safe landing execution. 

Our approach combines standard tracking MPCs for the multirotor \cite{nan2022nonlinear} and USV \cite{fossen2011handbook} and augments them with additional artificial goal locations. These artificial goals enable the vehicles to coordinate without prior guidance. Each vehicle solves an individual optimization problem for both the artificial goal and an input that tracks it, but only communicates the former to the other vehicle. The difference between the artificial multirotor and USV goals is penalized by a suitable coupling cost in both MPCs to enable consensus. 

Furthermore, our proposed distributed MPC integrates a spatial-temporal wave model. The wave model maps the location and time to the tilt of the USV; see one such model in \cite{sears2023mapping}. Our proposed distributed MPC simultaneously finds a consensus landing location between the UAV and USV (through a cooperation cost), tracks it (through a tracking cost) and optimizes a location and time that aids safe landing by minimizing large tilt angles of the USV (through a tilt cost). The two key contributions of this paper are:
\begin{itemize}
	\item We develop a novel distributed MPC framework for safe UAV-USV cooperative landing that leverages a spatial-temporal wave model. 
	\item We show in simulation how our approach can simultaneously coordinate in real-time both a safe landing location and execute the landing task for a UAV on a USV under wave conditions.  
\end{itemize}

\section{PROBLEM STATEMENT}
The goal is to achieve cooperation between the two agents to solve the problem of coordinated landing of a multirotor on a USV. More precisely, their outputs (i.e., the positions of the multirotor and USV) should converge to the set of equal output values, i.e.,
$$\lim_{t\rightarrow \infty} \mathbf{e}(t) = 0,$$
where $\mathbf{e}(t) = \mathbf{p}^m - \mathbf{p}^s$ is the error between the position of the multirotor $\mathbf{p}^m$ and the surface vessel $\mathbf{p}^s$. We consider a multi-agent heterogeneous system comprising two robotic agents, i.e., a multirotor and USV. The multirotor has nonlinear dynamics given by,
\begin{equation}
	\mathbf{\dot{x}}^m =f_M(\mathbf{x}^m, \mathbf{u}^m),
 \label{eq_uav_model}
\end{equation}
where $\mathbf{x}^m$ is the multirotor state, $\mathbf{u}^m$ is the multirotor input. The USV has nonlinear dynamics:
\begin{equation}
\mathbf{\dot{x}}^s =f_S(\mathbf{x}^s, \mathbf{u}^s),
\label{eq_usv_model}
\end{equation}
where $\mathbf{x}^s$ is the USV state, $\mathbf{u}^s$ is the USV input.
We assume that state measurements are available for both the multirotor and USV. We assume that communication is bilateral between the multirotor and surface vessel. This work aims to solve the autonomous landing and general consensus problem in challenging wave conditions while minimizing shared information, i.e., communication between vessels is limited and may be delayed. We propose a distributed Model Predictive Control (MPC) strategy to solve this problem.

\section{BACKGROUND}

\subsection{Multirotor Dynamics}

We model the multirotor as a rigid body in (\ref{eq_multirotor_dyn}). The dynamics of the multirotor system $\mathbf{\dot{x}}^m =f_M(\mathbf{x}^m, \mathbf{u}^m)$, taken from \cite{nan2022nonlinear}, can be written as:
\begin{equation}
    \begin{aligned}
 \mathbf{\dot{p}}^m &= \mathbf{v}^m, \\
\mathbf{\dot{v}}^m &= \mathbf{q}^m \odot \mathbf{c} - \mathbf{g},\\
\mathbf{\dot{q}}^m &= \frac{1}{2} \Lambda(\pmb{\omega}^m_B) \cdot \mathbf{q}^m,
\label{eq_multirotor_dyn}
\end{aligned}
\end{equation}
where $\mathbf{p}^m = [p_x^m, p_y^m, p_z^m]^T$ and $\mathbf{\dot{v}}^m= [v_x^m, v_y^m, v_z^m]^T$ are the position and the velocity vectors of the multirotor in the world frame $W$. We use a unit quaternion $\mathbf{q}^m=[q_w,q_x,q_y,q_z]^T$ to represent the orientation of the multirotor and $\pmb{\omega}^m_B = [\omega^m_x, \omega^m_y, \omega^m_z]^T$ to denote the body rates in the body frame $B$. Here, $g=[0,0,-g_z]^T$ with $g_z=9.81$ m/s$^2$ is the gravity vector, and $\Lambda(\pmb{\omega}^m_B)$ is a skew-symmetric matrix.  Finally, $\mathbf{c}=[0,0,c]^T$ is the mass-normalized thrust vector. We use a state vector $\mathbf{x}^m=[\mathbf{p}^m, \mathbf{v}^m, \mathbf{q}^m]^T$ and an input vector $\mathbf{u}^m = [c, \omega^m_x, \omega^m_y, \omega^m_z]^T$.

\subsection{Uncrewed Surface Vessel Dynamics}

We model the marine craft surface vessel $\mathbf{\dot{x}}^s =f_S(\mathbf{x}^s, \mathbf{u}^s)$ using the rigid-body dynamics for the Maritime Robotics Otter USV in \cite{fossen2011handbook} as:
\begin{equation}
    \mathbf{M} \pmb{\dot{\nu}}^s + \mathbf{C}(\pmb{\nu}^s) \pmb{\dot{\nu}}^s + \mathbf{D}(\pmb{\nu}^s) \pmb{\dot{\nu}}^s + g(\pmb{\eta}^s) + \mathbf{g}_0 = \pmb{\tau},
    \label{usv_model}
\end{equation}

where $\pmb{\nu}^s = [v_x^s, v_y^s, v_z^s, p, q, r]^T$ and $\pmb{\eta}^s = [p_x^s, p_y^s, p_z^s, \alpha, \beta, \psi]^T$ are generalized velocities and positions in the world frame used to describe the surface vessel motions in six degrees of freedom and $\pmb{\tau} \in \mathbb{R}^6$ are the generalized forces acting on the craft. In this model $\mathbf{M}$, $\mathbf{C}(\pmb{\nu}^s)$ and $\mathbf{D}(\pmb{\nu}^s)$ denotes the inertia, Coriolis and damping matrices, $g(\pmb{\eta}^s)$ is the generalized gravitational and buoyancy force-matrix and $\mathbf{g}_0$ consists of static restoring forces and moments due to ballast systems and water tanks. We use a state vector $\mathbf{x}^s=[\pmb{\eta}^s, \pmb{\nu}^s]^T$ and an input vector $\mathbf{u}^s = \pmb{\tau}$.

\subsection{Spatial-Temporal Map of Waves}

Spatial-temporal maps are data-driven estimates of time-changing phenomena. The USV can be used as a mobile sensing platform to make observations about the waves. While this limits when and where data is collected, as demonstrated in \cite{sears2023mapping}, Gaussian Process (GP) regression can be used to create a spatial-temporal wave model by assuming spatial and temporal correlations in the data through kernel functions. When the USV crosses a wave at some point, there will be a measurable change in the vehicle's pitch $\alpha$ and roll $\beta$ angles. The wave or \textit{tilt} angle $\phi$ (the angle between the direction of the gravity vector measured in the USV body frame and the gravity vector measured in the inertial frame) is then:
\begin{equation}
	\phi = \arccos(\cos(\alpha) \cos(\beta)), 
	\label{eq_tilt}
\end{equation}
At any vessel position $x$ and $y$ and time $t$, inertial measurements can obtain an estimated pitch $\hat{\alpha}$ and estimated roll $\hat{\beta}$. Using (\ref{eq_tilt}), an estimated tilt angle $\hat{\phi}$ is determined. 

In this paper, we make use of a spatial-temporal wave map $f_W(\mathbf{x}) : \mathbb{R}^{\text{dim}(\mathbf{x})} \rightarrow \mathbb{R}$, from input $\mathbf{x} = [x, y, t]$ to the function output, i.e., tilt angle squared $\phi^2$ as:
\begin{equation}
   \phi^2 = f_W(\mathbf{x}).
   \label{eq_wave_1}
\end{equation}
One approach is to learn this model using GP regression. GP regression can  approximate the spatial-temporal wave map $f_W(\mathbf{x}) : \mathbb{R}^{\text{dim}(\mathbf{x})} \rightarrow \mathbb{R}$, from input $\mathbf{x} = [x, y, t]$ to the function output, i.e., tilt angle squared $\phi^2$. It does this by assuming that the function values $f_W(\mathbf{x})$, associated with different inputs $\mathbf{x}$, are random variables and that any finite number of these random variables have a joint Gaussian distribution. This nonparametric approach still requires us to define two priors: a prior mean function of $f_W(\mathbf{x})$, generally set to zero, and a covariance or kernel function $k(\cdot, \cdot)$ which encodes, for two input points, how similar their respective function values are. For the selection of a suitable kernel function for spatial-temporal wave mapping, see \cite{sears2023mapping}. 

This GP framework can be used to predict the function value, i.e., tilt angle, at any query point $\mathbf{x^*}$, i.e., at any vehicle position and time, based on $N$ noisy observations, $\mathcal{D} = \{ \mathbf{x}_i, \hat{\phi}^2_i \}^N_{i=1}$. The predicted mean and variance at the query point $\mathbf{x^*}$ conditioned on the observed data $\mathcal{D}$ are:
\begin{equation}
	\mu(\mathbf{x^*}) = \mathbf{k}(\mathbf{x^*}) \mathbf{K}^{-1} \mathbf{\hat{\Phi}}, 
	 \label{eq_mean}
\end{equation}
	 \vspace{-4mm}
\begin{equation}
	\sigma^2(\mathbf{x^*}) = k(\mathbf{x^*}, \mathbf{x^*}) - \mathbf{k}(\mathbf{x^*}) \mathbf{K}^{-1} \mathbf{k}^T(\mathbf{x^*}),
	\label{eq_var}
\end{equation}
\noindent
where $\mathbf{\hat{\Phi}} =[\hat{\phi}^2_1, \hat{\phi}^2_2, ..., \hat{\phi}^2_N]^T$ is the vector of observed function values, the covariance matrix has entries $\mathbf{K}_{(i,j)} = k(\mathbf{x}_i, \mathbf{x}_j), \quad i,j \in {1, ..., N}$, and $\mathbf{k}(\mathbf{x}) = [k(\mathbf{x^*}, \mathbf{x}_1), ..., k(\mathbf{x^*}, \mathbf{x}_N)]$ is the vector of the covariances between the query point $\mathbf{x^*}$ and the observed data points in $\mathcal{D}$. In this way, the USV acts as a `sensor' that can be used to model the wave map $f_W(\mathbf{x})$ as it moves in space and time. In this work, we will assume that the spatial-temporal wave model (\ref{eq_wave_1}) is known and show how it can be exploited to coordinate in real-time a location and time to land a multirotor on a USV. Future work will explore how the online learning of this model (\ref{eq_wave_1}) through GP regression impacts the coordinated landing strategy presented in this paper. 

\section{METHODOLOGY}
\subsection{Distributed Model Predictive Control}
Our proposed approach leverages a sequential distributed model predictive control (MPC) scheme for cooperative control of multi-agent systems with dynamically decoupled heterogeneous nonlinear agents subject to individual constraints.  Specifically, we introduce an artificial goal in tracking MPC for both the USV and multirotor. Each vehicle solves an individual optimization problem for an artificial goal and an input that tracks it, only communicating the former. In other words, the proposed scheme comprises two parts. In the tracking part, the nonlinear dynamics and constraints of the individual robots are handled in MPC. An additional designed cost augments the MPC for both robots. This term penalizes the deviation of the artificial outputs, i.e., to encourage a cooperative landing goal location, and the predicted tilt angle squared $\phi^2$ at the artificial outputs to promote the selection of a location and timing of a suitable landing goal for a safe, smooth landing. 

\paragraph{Uncrewed Surface Vessel MPC}

In MPC, we approximate the actual continuous-time differential equation using discrete-time integration, e.g., $\mathbf{x}^s_{i+1} = \mathbf{x}^s_{i} + \delta_s f_S(\mathbf{x}^s_{i}, \mathbf{u}^s_{i})$, with $\delta_s$ as the time interval between consecutive states of the USV and $f_S$ is the USV dynamics model in (\ref{eq_usv_model}).

The USV MPC takes the current state $\mathbf{x}^s_{i} = \mathbf{x}^s_{\text{init}}$ at each time step $i$. MPC produces a sequence of optimal system states $\mathbf{x}^{s*}_{0:N_S}$ and control commands $\mathbf{u}^{s*}_{0:N_S-1}$ by solving an optimization online using a multiple-shooting scheme. The notation $*$ denotes the optimal solution and $_{0:N_S}$ denotes the value for each time step from the current time step $i$ to $i+N_S$ where $N_S \in \mathbb{Z}$ is the prediction horizon. The first control command is applied to the surface vessel (\ref{usv_model}), after which the optimization problem is solved again in the next state. MPC requires minimizing a cost $J_S(\cdot)$ over a fixed time horizon $N_S$ at each control time step by solving a constrained optimization:
\begin{equation}
\begin{aligned}
	\min_{\mathbf{x}^s_{0:N_S}, \mathbf{u}^s_{0:N_S-1}, \mathbf{p}^s_G} & \quad J_S(\mathbf{x}^s_{0:N_S}, \mathbf{u}^s_{0:N_S-1}, \mathbf{p}^s_G, \mathbf{p}^m_G) \\
\textrm{s.t.} \quad & \mathbf{x}^s_{k+1} = \mathbf{x}^s_{k} + \delta_s f_S(\mathbf{x}^s_{k}, \mathbf{u}^s_{k}) \quad \forall k \in \mathcal{K}_s \\ 
  &\mathbf{x}^s_{k} \in \mathbb{X}_s \quad \forall k \in \mathcal{K}_s  \\
  &\mathbf{u}^s_{k} \in \mathbb{U}_s \quad  \forall k \in \mathcal{K}_s  \\
  &\mathbf{x}^s_{0} = \mathbf{x}^s_{\text{init}}
\end{aligned}
\label{eq_mpc_s}
\end{equation}
where $\mathcal{K}_s := \mathbb{Z} \cap [0, N_S-1]$, $\mathcal{X}_s$ are constraints on the USV state and $\mathcal{U}_s$ are constraints on the USV input. Unlike a standard tracking MPC in \cite{fossen2011handbook}, we also optimize for the artificial USV position goal $\mathbf{p}^s_G$ at each time step, where the last artificial multirotor goal $\mathbf{p}^m_G$ is communicated to the USV. The artificial multirotor goal $\mathbf{p}^m_G$ is sequentially optimized in the multirotor MPC using the latest communicated USV goal $\mathbf{p}^s_G$. As illustrated in Fig. 1, the USV and multirotor do not communicate their planned trajectories or current state to each other. Our proposed distributed MPC scheme only requires each vehicle to update and communicate its respective position goal. This makes communication relatively lightweight in contrast to alternative distributed MPC approaches in \cite{bereza2020distributed}.  

\paragraph{Multirotor MPC}
Similar to USV MPC, we use a discretization at time step $j$, $\mathbf{x}^m_{j+1} = \mathbf{x}^m_{j} + \delta_m f_M(\mathbf{x}^m_{j}, \mathbf{u}^m_{j})$, with $\delta_m$ as the time interval between consecutive states of the multirotor where $f_M(\mathbf{x}^m_{j})$ is the multirotor dynamics model in (\ref{eq_uav_model}). The multirotor MPC takes the current state $\mathbf{x}^s_{j} = \mathbf{x}^s_{\text{init}}$ at each time step $j$. MPC produces a sequence of optimal system states $\mathbf{x}^{m*}_{0:N_M}$ and control commands $\mathbf{u}^{m*}_{0:N_M-1}$ by solving an optimization online using a multiple-shooting scheme. The notation $*$ denotes the optimal solution and $_{0:N_M}$ denotes the value for each time step from the current time step $j$ to $j+N_M$ where $N_M \in \mathbb{Z}$ is the prediction horizon. The multirotor MPC minimizes a cost $J_M(\cdot)$ over a fixed time horizon $N_M$ at each control time step by solving a constrained optimization:
\begin{equation}
\begin{aligned}
	\min_{\mathbf{x}^m_{0:N_M}, \mathbf{u}^m_{0:N_M-1}, \mathbf{p}^m_G} & \quad J_M(\mathbf{x}^m_{0:N_M}, \mathbf{u}^m_{0:N_M-1}, \mathbf{p}^m_G, \mathbf{p}^s_G) \\
\textrm{s.t.} \quad & \mathbf{x}^m_{k+1} = \mathbf{x}^m_{k} + \delta_m f_M(\mathbf{x}^m_{k}, \mathbf{u}^m_{k}) \quad \forall k \in \mathcal{K}_m\\
  &\mathbf{x}^m_{k} \in \mathcal{X}_m \quad \forall k \in \mathcal{K}_m \\
  &\mathbf{u}^m_{k} \in \mathcal{U}_m \quad  \forall k \in \mathcal{K}_m  \\
  &\mathbf{x}^m_{0} = \mathbf{x}^m_{\text{init}}
\end{aligned}
\label{eq_mpc_m}
\end{equation}
where $\mathcal{K}_m := \mathbb{Z} \cap [0, N_M-1]$, $\mathcal{X}_m$ are constraints on the multirotor state and $\mathcal{U}_m$ are constraints on the multirotor input. Unlike a standard tracking MPC in \cite{nan2022nonlinear}, we also optimize for the artificial multirotor position goal $\mathbf{p}^m_G$ at each time step, where the last artificial USV goal $\mathbf{p}^s_G$ is communicated from the USV. The artificial USV goal $\mathbf{p}^s_G$ is simultaneously sequentially optimized in the USV MPC in (\ref{eq_mpc_s}). 

\subsection{MPC Objective Functions}

We propose a novel cost for the MPC multirotor and USV controllers:
\begin{equation}
    J_M(\cdot) = J^{\textrm{track}}_M(\cdot) + J_M^{\textrm{co-op}}(\cdot) + J^{\textrm{tilt}}_M(\cdot)
    \label{eq_cost_M}
\end{equation}
\begin{equation}
   J_S(\cdot) = J^{\textrm{track}}_S(\cdot) + J_S^{\textrm{co-op}}(\cdot) + J^{\textrm{tilt}}_S(\cdot) 
   \label{eq_cost_S}
\end{equation}
where $J^{\textrm{track}}_S(\cdot)$ is a standard quadratic tracking cost for the surface vessel, $J^{\textrm{track}}_M(\cdot)$ is a standard quadratic tracking cost for the multirotor. Specifically the tracking cost for the multirotor $J^{\textrm{track}}_M(\cdot)$ is given as:
\begin{equation}
    \sum_{k=1}^{N_M} (\mathbf{C} \mathbf{{x}}^m_k - \mathbf{p}^m_G)^T\mathbf{Q}^m(\mathbf{C} \mathbf{{x}}^m_k - \mathbf{p}^m_G) + {\mathbf{u}^m_{k-1}}^T \mathbf{R}^m \mathbf{u}^m_{k-1}
    \label{eq_cost_M_track}
\end{equation}
where the $\mathbf{p}^m_k = \mathbf{C} \mathbf{{x}}^m_k$ is the position of the multirotor at step $k$, $\mathbf{Q}^m \succ 0$ and $\mathbf{R}^m \succ 0$ are selected positive definite matrices that weight the position error (between the multirotor position and its goal) and control effort respectively. Similarly, the tracking cost for the surface vessel $J^{\textrm{track}}_S(\cdot)$ is given as:
\begin{equation}
    \sum_{k=1}^{N_S} (\mathbf{C} \mathbf{{x}}^s_k - \mathbf{p}^s_G)^T\mathbf{Q}^s(\mathbf{C} \mathbf{{x}}^s_k - \mathbf{p}^s_G) + {\mathbf{u}^s_{k-1}}^T \mathbf{R}^s \mathbf{u}^s_{k-1},
    \label{eq_cost_S_track}
\end{equation}
where the $\mathbf{p}^s_k = \mathbf{C} \mathbf{{x}}^s_k$ is the position of the multirotor at step $k$, $\mathbf{Q}^s \succ 0$ and $\mathbf{R}^s \succ 0$ are selected positive definite matrices that weight the position error (between the surface vessel position and its goal) and control effort respectively.

In the MPC cost for both the multirotor (\ref{eq_cost_M}) and surface vessel (\ref{eq_cost_S}), we include a cooperation cost. At each time step, the multirotor receives the surface vessel's current goal $\mathbf{p}_{G}^s$ and aims to optimize its goal location $\mathbf{p}_{G}^m$ to cooperate with the surface vessel through a cooperation cost:
\begin{equation}
    J_M^{\textrm{co-op}}(\cdot) = (\mathbf{p}^m_G - \mathbf{p}^s_G - h_d \mathbf{e}_3)^T \mathbf{W}^m (\mathbf{p}^m_G - \mathbf{p}^s_G - h_d \mathbf{e}_3),
    \label{eq_cost_coop_M}
\end{equation}
where $\mathbf{e}_3 = [0, 0, 1]^T$ and $\mathbf{W}^m = \text{diag}(w^m_x, w^m_y, w^m_z) \succ 0$ weights the error in the goal location of the multirotor $\mathbf{p}^m_G$ (which is optimized for at each time step in (\ref{eq_mpc_m})), and the surface vessel's current goal $\mathbf{p}_{G}^s$, communicated to the multirotor from the surface vessel as shown in Fig. 1, as well as error between the height of the multirotor's goal $z^{m}_G$ and surface vessel's goal $z^{s}_G$ plus a holding height $h_d$. This encourages the multirotor's goal to converge to the same $x-y$ location as the surface vessel with a height of $h_d$ above the vessel. Furthermore, it is unsafe for the multirotor to have a height below the surface vessel in its descent (likely resulting in a crash). To enforce this we will first define the error $\mathbf{e}_{G,z}$ as the difference in height between the multirotor's goal $z^{m}_G$ and surface vessel's goal $z^{s}_G$:
\begin{equation}
    \mathbf{e}_{G,z} = z^{m}_G - z^{s}_G.
\end{equation}
and $\mathbf{e}_{{G,z}_k}$ is the height error at time step $k$. We impose additional constraints (to ensure that the multirotor does not go below the USV) in the multirotor MPC (\ref{eq_mpc_m}):
\begin{equation}
    \mathbf{e}_{{G,z}_k} \geq 0 \quad \forall k=0, ..., N_M,
\end{equation}
\begin{equation}
    z^m_k - z^{m}_G \geq 0 \quad \forall k=0, ..., N_M,
\end{equation}
where $z^m_k$ is the height of the multirotor at time step $k$.

We consider a similar cooperation cost in the surface vessel cost (\ref{eq_cost_S}). Each time the MPC is performed, the surface vessel receives the multirotor's current goal $\mathbf{p}_{G}^m$ and aims to optimize its goal location $\mathbf{p}_{G}^s$ to cooperate with the multirotor through a cooperation cost:
\begin{equation}
    J_S^{\textrm{co-op}}(\cdot) = (\mathbf{p}^s_G - \mathbf{p}^m_G)^T \mathbf{W}^s (\mathbf{p}^s_G - \mathbf{p}^m_G),
    \label{eq_cost_coop_S}
\end{equation}
where diagonal matrix $\mathbf{W}^s = \text{diag}(w^s_x, w^s_y, 0)$, $w^s_x \geq 0$ and $w^s_y \geq 0$, weights the error in the goal $x-y$ location of the surface vessel $\mathbf{p}^s_G$ (which is optimized for at each time step in (\ref{eq_mpc_s})), and the multirotor's current goal $\mathbf{p}_{G}^m$, communicated to the surface vessel from the multirotor as shown in Fig. 1.

To ensure a safe and soft landing, we would also like to select a spatial location with relatively small waves or ``calmer waters" (i.e., a small tilt angle on average). To do this, we include a cost for the surface vessel $J^{\textrm{tilt}}_S(\cdot)$ that aims to select a goal location $\mathbf{p}^s_G = [x^s_G, y^s_G, z^s_G]^T$ where the average squared tilt angle of a wave over time (with time period $\delta_w N_W$) is minimized as:
\begin{equation}
    J^{\textrm{tilt}}_S(\cdot) = \lambda_s \frac{\sum_{j=0}^{N_W} f_W([x^s_G, y^s_G, \delta_w j]^T)}{N_W + 1},
    \label{eq_tilt_S}
\end{equation}
where $f_W$ is the spatial-temporal wave model that outputs the squared tilt angle $\phi^2$ in (\ref{eq_wave_1}) as a function of input $\mathbf{x} = [x, y, t]^T$ comprising of the spatial location $(x, y)$ and the time $t$, $\delta_w$ is the discretization interval of the wave, $\delta_w N_W$ is the time period such that $f_W([x^s_G, y^s_G, 0]^T) = f_W([x^s_G, y^s_G, \delta_w N_W]^T)$. The cost is used to drive the surface vessel's goal location $\mathbf{p}^s_G$ to a location of calmer waters while encouraging cooperation with the multirotor for landing. The weight $\lambda_s>0$ can be increased to drive the surface vessel's goal location towards calmer water first before cooperating a landing location with the multirotor.  

Our proposed MPC optimizes a safe (small tilt) landing location by including the cost $J^{\textrm{tilt}}_S(\cdot)$ for the surface vessel. We include the cost $J^{\textrm{tilt}}_M(\cdot)$ for the multirotor to time the landing of the multirotor on the USV when its tilt is low. To do this, we adopt an approach similar to \cite{gupta2022landing} where we activate or include $J^{\textrm{tilt}}_M(\cdot)$ in the MPC cost (\ref{eq_cost_M}) only when certain conditions hold. Three conditions must hold to include $J^{\textrm{tilt}}_M(\cdot)$. The first condition is that the surface vessel is at its current goal within some small threshold $\epsilon_1 \geq 0$, i.e., 
\begin{equation}
    | \mathbf{p}^s - \mathbf{p}^s_G | \leq \epsilon_1.
    \label{eq_cond1}
\end{equation}
When this happens, the surface vessel communicates to the multirotor that ``landing is possible". The multirotor then checks two more conditions. The second condition is that the multirotor's current goal is within some threshold $\epsilon_2 \geq 0$ of the surface vessel's current goal:
\begin{equation}
    | \mathbf{p}_G^m - \mathbf{p}^s_G | \leq \epsilon_2.
    \label{eq_cond2}
\end{equation}
The final condition is that the multirotor's current position is within some threshold $\epsilon_3 \geq 0$ of its goal: 
\begin{equation}
    | \mathbf{p}^m - \mathbf{p}^m_G | \leq \epsilon_3.
    \label{eq_cond3}
\end{equation}
When conditions (\ref{eq_cond1})-(\ref{eq_cond3}) hold, then $J^{\textrm{tilt}}_M(\cdot)$ is activated as:
\begin{equation}
    J^{\textrm{tilt}}_M(\cdot) = \lambda_m \sum_{k=1}^{N} h(\mathbf{e}_{G,z}) f_W([x^s_G, y^s_G, t_k]^T),
    \label{eq_tilt_M}
\end{equation}
where $t_k = t_c + \delta_m k$, $t_c$ is the current time each time the OCP is solved in the MPC in (\ref{eq_mpc_m}), $\delta_m$ is discretization of the multirotor model used in (\ref{eq_mpc_m}), $\lambda_M > 0$ is a user-selected weight and $f_W$ is the wave model in (\ref{eq_wave_1}). This term considers the squared tilt angle $\phi^2$ at the surface vessel's goal location, which is sent from the surface vessel and fixed with respect to the multirotor's objective function (\ref{eq_cost_M}). 
The landing function $h(\mathbf{e}_{G,z})$ is selected similar to that in \cite{gupta2022landing} as:
\begin{equation}
    h(\mathbf{e}_{G,z}) = \begin{cases}
(1 + \text{exp}(-\frac{\mathbf{e}_{G,z} - h_d}{-0.15}))^{-1} &\text{, if $\mathbf{e}_{G,z} \geq 0.16$}\\
(1 + \text{exp}(\frac{\mathbf{e}_{G,z} - h_d}{-0.01}))^{-1} &\text{, otherwise}
\end{cases}
\label{eq_land_fcn}
\end{equation}

\begin{figure}[H]
 \centering
 \subfigure[``Cooperative" Strategy]{
      \centering
      \label{strategy_coop}
      \includegraphics[width=0.35\textwidth, trim={0cm 0cm 1cm 1cm},clip]{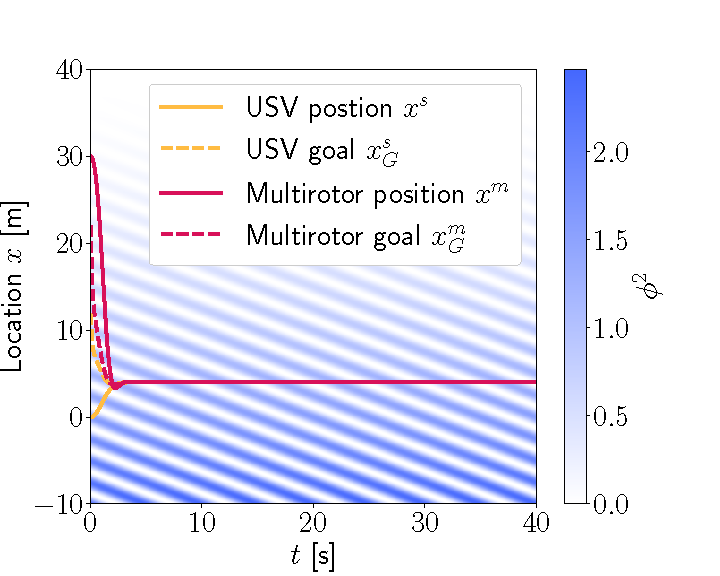}
		} 
 \subfigure[``Calm" Strategy]{
      \centering
      \label{strategy_calm}
      \includegraphics[width=0.35\textwidth, trim={0cm 0cm 1cm 1cm},clip]{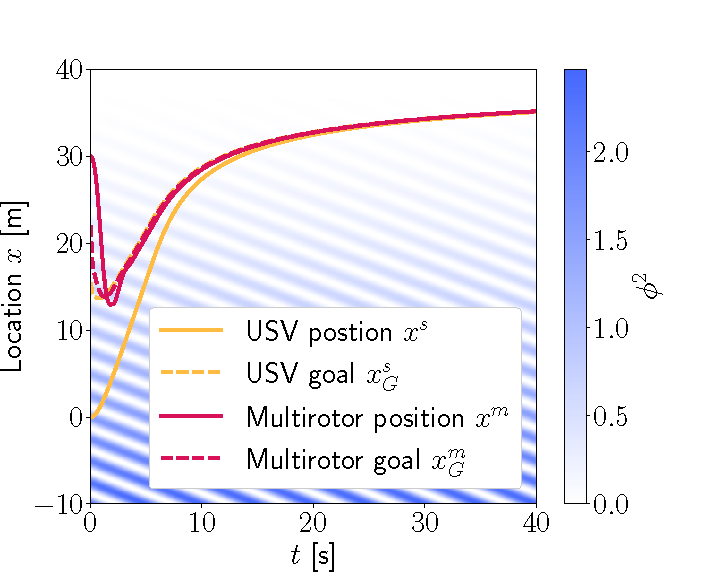}
		} 
\subfigure[``Ride the Wave" Strategy]{
      \centering
      \label{strategy_ride}
      \includegraphics[width=0.35\textwidth, trim={0cm 0cm 1cm 1cm},clip]{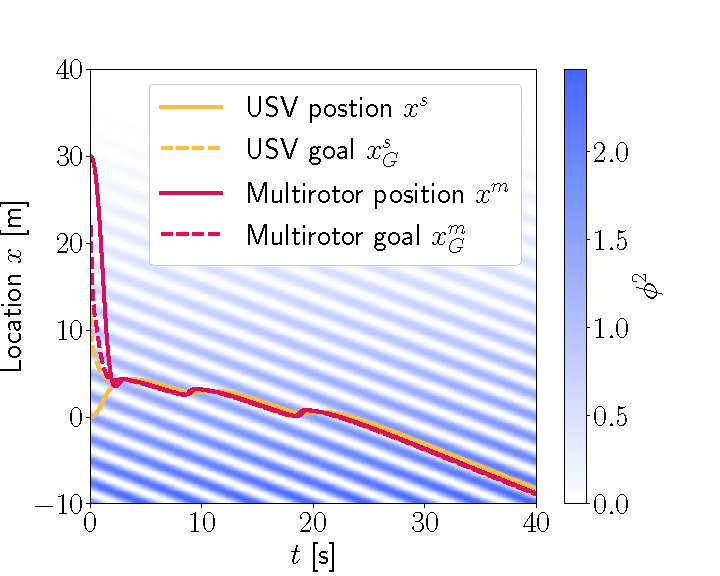}
		} 
 \caption{Visualization of multirotor trajectory (solid red), multirotor goal (dashed red), USV trajectory (solid yellow) and USV goal (dashed yellow) using a distributed MPC framework with no tilt cost $J^{\text{tilt}}(\cdot)$ for either vehicle in (a) ``Cooperative" Strategy, our proposed $J^{\text{tilt}}_S(\cdot)$ in (b) ``Calm" Strategy and an alternative tilt cost in (c) ``Ride the Wave" Strategy. Our proposed ``Calm" Strategy leads to a lower tilt of the USV before reaching a consensus on the final landing location.} 
 \label{fig_strategy_results}
\end{figure}

\noindent
where $h_d$ is the holding location height during a landing attempt.

\section{SIMULATION RESULTS}
We perform 2D simulation experiments where we implement each MPC in a thread. We augment existing tracking MPCs for the USV \cite{fossen2011handbook} and multirotor \cite{song2022policy}. Both controllers have a look-ahead time of 2 seconds. The multirotor MPC operates at a frequency of 50Hz, while the USV MPC operates at 10Hz. We set the following parameters: simulation time of 60 seconds, initial position vectors $\mathbf{p}^m = [30, 0, 5]^T$ and $\pmb{\eta}^s = [0,0,0,0,0,0]^T$, and initial artificial random positional goals $\mathbf{p}_{G}^m=[35,0,0]$ and $\mathbf{p}_{G}^s=[20,0]$. Weight matrices in the tracking and cooperation costs are selected as $\mathbf{Q}^s=\text{diag}(1000,1000)$, $\mathbf{R}^s=\text{diag}(0.1,0.1)$, $\mathbf{W}^s=\text{diag}(1000,0)$, $\mathbf{Q}^m=\text{diag}(1000,0,100)$, $\mathbf{R}^m=\text{diag}(0.1,0.1,0.1,0.1)$, and $\mathbf{W}^m=\text{diag}(1000, 100)$. The amplitude of the wave $A_W=-\frac{1}{20}(x - 40)\sin\left(\frac{\pi}{5}(t+2x)\right)$ where the spatial-temporal wave model $f_W=(\frac{\partial A_w}{\partial t})^{2}$.

\begin{figure}
\centering
\includegraphics[width=0.4\textwidth, trim={0cm 0cm 0 0cm},clip]{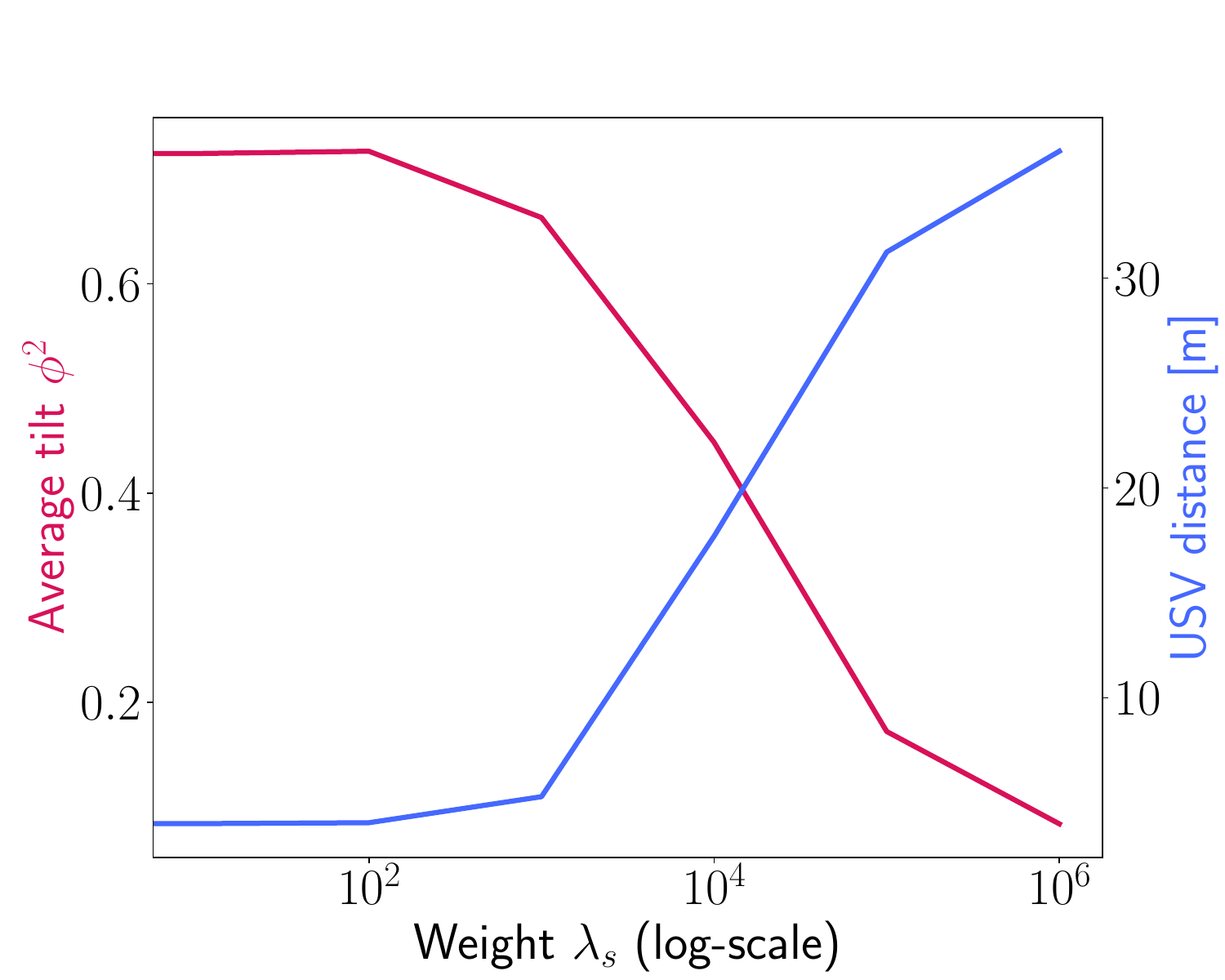}
\caption{Trade-off between average tilt at the landing location vs the distance the USV travels. As we increase the weight on $\lambda_S$, the tilt at landing decreases but the USV has to travel further to enable this.}
\label{fig_weight}
\end{figure}

\subsection{Spatial Cooperation}

We visualize the spatial consensus of the proposed distributed MPC scheme under three strategies in Fig. \ref{fig_strategy_results}. These three strategies are: the ``Cooperative" strategy, no tilt cost $J^{\text{tilt}}(\cdot)$ for either vehicle is considered; our proposed ``Calm" strategy that leverages the cost (\ref{eq_tilt_S}) with $\lambda_s = 10^5$; and an alternative ``Ride the Wave" strategy. In the ``Ride the Wave" strategy, the tilt cost (\ref{eq_tilt_S}) is replaced with a short-term tilt cost $\sum_{k=1}^{N_S} f_W([x^s_G, y^s_G, t_k]^T)$. A purely ``Cooperative" strategy reaches a consensus location for landing relatively quickly (at around $5$ s) in Fig. \ref{fig_strategy_results}. However, as we observe in Fig. \ref{fig_weight}, the short distance and time to landing do not account for the USV tilt at the landing location leading to a potentially unsafe landing on large waves. The ``Ride the Wave" strategy reaches consensus at around $4.9$ m, similar to the ``Cooperative" strategy. This position is maintained until the first trough of the wave passes the momentarily stationary USV. When the vessel is in a trough, $\phi^2$ is minimized, triggering the USV to follow the trough of the wave. If the USV moves too close to one side of the wave, as it does twice in Fig. \ref{strategy_ride}, the trough of the adjacent wave may be perceived in the prediction horizon encouraging the USV to move across a peak to reach that adjacent trough. The multirotor (solid red) and USV (solid yellow) do not converge because the USV goal (dashed yellow) is continuously changing, i.e., riding the wave. While an interesting strategy for non-stationary multirotor-USV landing, a safer approach (``Calm" strategy) is proposed in this paper. Our proposed ``Calm" strategy, see Fig. \ref{strategy_calm}, drives the USV goal towards calmer waters, converging at $\sim 36$ m. Initially, the trajectories are similar to the ``Cooperative" strategy. However, in the ``Calm" strategy, the USV goal shifts towards the location of decreasing average tilt $\phi^2$. Once the tilt cost is minimized, the USV resumes cooperation with the multirotor. 

\begin{figure}
 \centering
 \subfigure[Lost Communication at $0.02$ s]{
      \centering
      \includegraphics[width=0.35\textwidth, trim={0cm 0cm 1cm 1cm},clip]{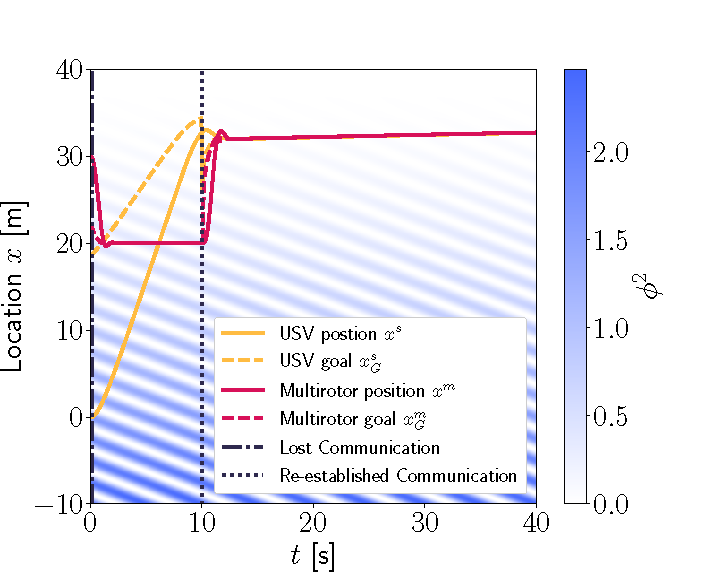}
      \label{comms_1}
		} 
 \subfigure[Lost Communication at $5$ s]{
      \centering
      \includegraphics[width=0.35\textwidth, trim={0cm 0cm 1cm 1cm},clip]{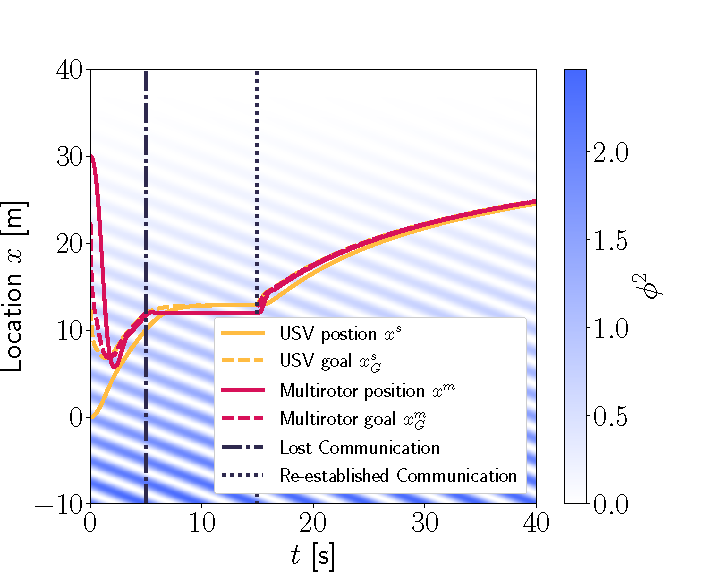}
      \label{comms_2}
		} 
 \caption{Visualization of multirotor trajectory (solid red), multirotor goal (dashed red), USV trajectory (solid yellow) and USV goal (dashed yellow) using a distributed MPC framework our proposed $J^{\text{tilt}}_S(\cdot)$ where there is a $10$ s communication loss between the vehicles at (a) $0.02$ s and (b) $5$ s. Our proposed approach is robust to communication breaks.} 
 \label{fig_comms}
\end{figure}

We illustrate the robustness of our distributed MPC to temporary communication losses in Fig. \ref{fig_comms} where both vehicles can not exchange their goals for $10$ s. We observe that each vehicle moves towards the last known goal of the other vehicle. When communication is re-established after $10$ s, the goals rapidly adjust to recalibrate the MPC consensus.

\subsection{Temporal Cooperation}
\begin{figure}
\centering
\includegraphics[width=0.4\textwidth, trim={0cm 0cm 0 0cm},clip]{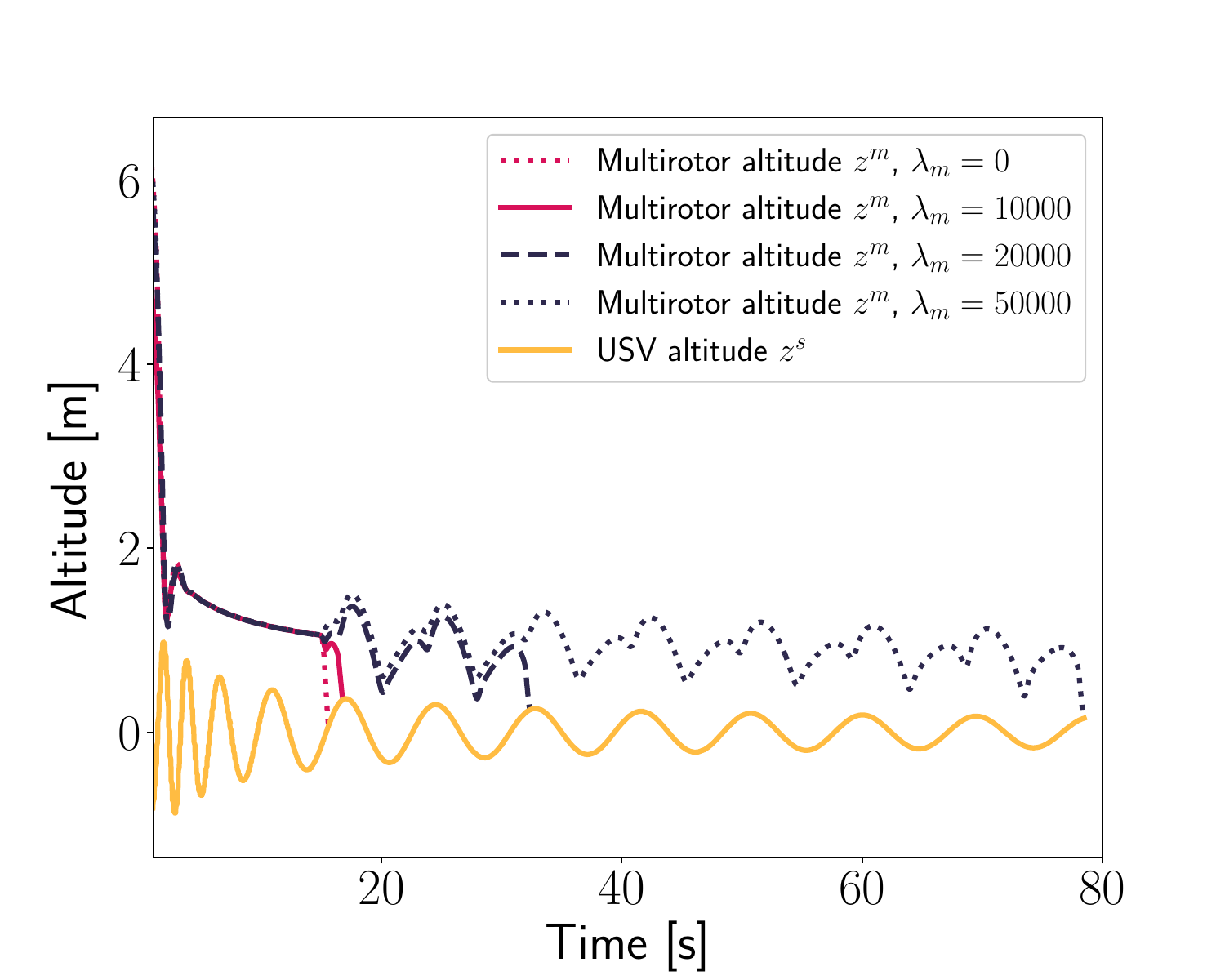}
\caption{Visualization of multirotor altitude as it descends to land for increasing $\lambda_m$. For $\lambda_m > 10000$ (dashed and dotted black), the multirotor is very sensitive to small changes in the tilt at landing and, therefore, takes longer to land. We propose $\lambda_m \approx 10000$ (solid red) to balance this trade-off. }
\label{fig_z}
\end{figure}

\begin{figure}
\centering
\includegraphics[width=0.4\textwidth, trim={0cm 0cm 0 0cm},clip]{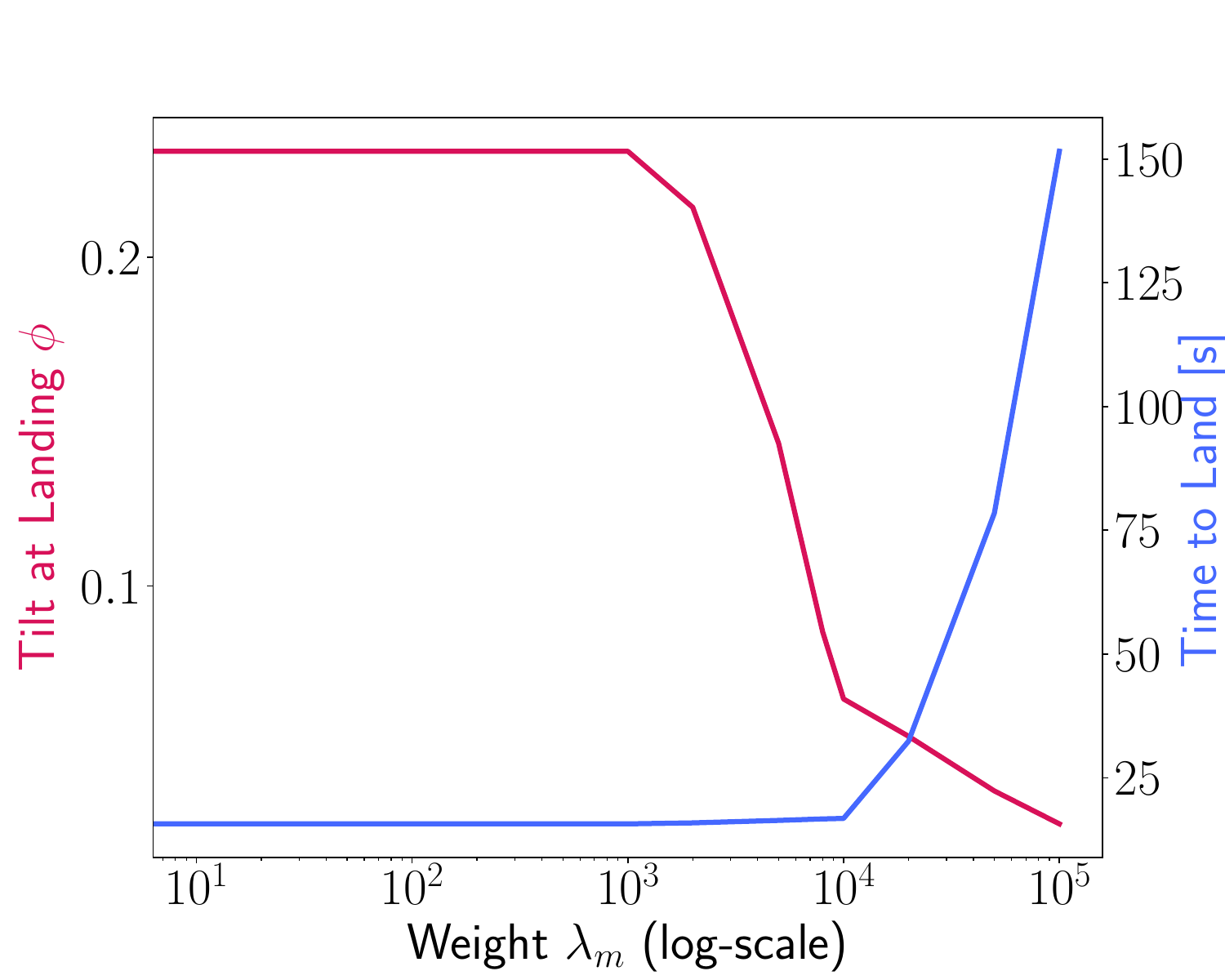}
\caption{Trade-off between tilt at the landing time vs the time for multirotor to land. As we increase the weight on $\lambda_m$, the tilt at landing decreases but the time to land increases.}

\label{fig_weight_m}
\end{figure}

 Fig. \ref{fig_z} shows the altitude of the multirotor as it descends from an initial height of $5$ m. We select $h_d = 1$ m. The tilt cost (\ref{eq_tilt_M}) is automatically applied at $\sim 15$ s when the landing conditions are met. We assume that the USV height changes with the wave. For $\lambda_m > 10000$ (dashed and dotted black), the multirotor is very sensitive to small changes in the tilt at landing and, therefore, takes longer to land. There is a trade-off between tilt at the landing time vs the time for multirotor to land, see Fig. \ref{fig_weight_m}. As we increase the weight on $\lambda_m$, the tilt at landing decreases, but the time to land increases. Beyond $\lambda_m = 10^4$, the decrease in tilt at landing is small compared to the increase in time to land. Our proposed approach achieves spatial-temporal cooperation for a multirotor landing on a USV by leveraging the USV to travel to ``Calm" waters and the multirotor to time the landing. 

 

\section{Conclusion}
This paper presents a novel distributed MPC strategy for the cooperative landing of a multirotor on a USV by leveraging a spatial-temporal wave model. We illustrate a method that can easily augment existing tracking MPC techniques for multirotors and USVs. We demonstrate its robustness to potential communication losses and delays. Significantly, our approach can coordinate and execute in real-time a safe landing location and time. In this work, we assumed that the wave model was known. In future work, we will explore the impact of learning this model online on the cooperative landing strategy presented and extend the results through experimental validation on a physical multirotor and USV system. 

\addtolength{\textheight}{-6cm}

\bibliography{References.bib}
\bibliographystyle{ieeetr}

\end{document}